\documentclass[runningheads]{llncs}
\usepackage[T1]{fontenc}
\usepackage{booktabs}   % in preamble
\usepackage{multirow}   % if needed
\usepackage{graphicx}
\usepackage{amsmath}
\usepackage{amssymb}
\usepackage{subcaption}
\usepackage{arydshln}  % in preamble
\usepackage{wrapfig}
\usepackage{colortbl}
\usepackage{xcolor}
\usepackage[colorlinks=true, linkcolor=blue, citecolor=blue, urlcolor=blue]{hyperref}
\usepackage{marvosym}

\usepackage{orcidlink}
\usepackage{fontawesome5}  % for ORCID icon

\begin{document}
\title{Toward Mask Annotation-Free Surgical Instrument Segmentation from Endoscopic Images Using Text-Prompted Segment Anything Model 3 
}
    \titlerunning{Mask Annotation-Free Instrument Segmentation Using SAM3}

\author{Nakul Poudel\inst{1}\textsuperscript{(\Letter)}\orcidlink{0009-0000-8957-9730} \and Richard Simon \inst{2} \and Cristian A. Linte\inst{1,2}\orcidlink{0000-0001-7602-7937}}

\authorrunning{N. Poudel {\it et al}.}

\institute{Center for Imaging Science, Rochester Institute of Technology, Rochester, NY 14623, USA \\
\and
Biomedical Engineering, Rochester Institute of Technology, Rochester, NY 14623, USA
\\
\email{\{np1140, rasbme, calbme\}}@rit.edu}

\maketitle              % typeset the header of the contribution
\begin{abstract}
Surgical instrument segmentation is a fundamental task for computer-assisted interventions, yet most existing methods rely on pixel-level annotations or manual spatial prompts, which limit scalability and automation. The recently introduced Segment Anything Model 3 (SAM3) offers a pathway to annotation-free, automatic segmentation via text-based prompting; however, the instrument name as a text prompt could not be directly used due to a large domain gap. To overcome these limitations, we propose a two-stage framework that achieves instance-level segmentation without requiring ground truth masks or manual interaction. In the first stage, we leverage a natural-language-aligned generic prompt -- "tool" -- to produce binary masks using SAM3’s zero-shot capability. In the second stage, these masks are extended to instance-level by integrating a vision–language model (Qwen) that is fine-tuned on SAM3-generated masked regions for instrument classification. We evaluate our approach on the EndoVis 2017 and 2018 datasets. Results show that, while our two-stage approach does not reach the performance of current fully supervised methods, it significantly outperforms the direct use of SAM3 for instance-level instrument segmentation with text prompts. Overall, our findings highlight both the limitations and potential of SAM3, suggesting a promising direction toward annotation-free surgical instrument segmentation.

\keywords{Surgical Instrument Segmentation  \and Image-Guided Surgery \and Segment Anything Model 3 (SAM3) \and Zero-Shot Segmentation}
\end{abstract}
\section{Introduction}

% Importance of surgical instrument segmentation. 
% Current problem in segmentation, 
% require large annotation data, and annotation is costly.
% Emergence of SAM, SAM provides zero shot segmentation. However requires manual interaction, limiting automation.
% Sam3 introduces text promptable thus avoiding the spatial interaction need. However as presented in table 1, when using instrument names the performance is very low compare to using spatial prompts (bboxes, and points). So, in this research we designed a framework leveraing sam3 using general and natural language aligned fixed text prompt "tool" to first extract binary segmentation masks, and later fine-tuned qwen model using sam3 generated combining sam3 generated mask with rgb image to classfy surgical intruments, thus geneerating complete instance level instrument segmentation without the requirement of grounth segmentationm masks. 

Surgical instrument segmentation, which identifies the precise spatial location of instruments in a surgical scene, is vital for developing computer-assisted surgical systems~\cite{ahmed2024deep}. Accurate segmentation supports several downstream surgical tasks, including surgical phase recognition, skill assessment, tool tracking, and pose estimation. However, the field mostly relies on fully supervised deep learning models that typically require large amounts of pixel-level annotated data. Creating such annotations requires specialized domain expertise and is labor-intensive and costly.

The Segment Anything Model (SAM)~\cite{kirillov2023segment} is a powerful zero-shot segmentation model with the potential to reduce reliance on manual annotations. However, SAM and its successor, SAM2~\cite{ravi2024sam}, depend on precise spatial prompts (e.g., points or bounding boxes) to guide segmentation. This requirement introduces human-in-the-loop interaction, which can limit automation and reduce the applicability of the models in surgical environments. Recently, SAM’s successor, SAM3~\cite{carion2025sam}, was introduced, supporting text-based prompting and potentially enabling segmentation without manual spatial inputs. However, directly applying SAM3 with instrument names as text prompts in surgical scenes remains challenging. Surgical instruments often exhibit similar textures, metallic reflections, and frequent occlusions, making semantic differentiation difficult. Furthermore, as SAM3 is primarily trained on natural images and associated text, a significant domain gap arises, limiting its ability to effectively handle instrument-specific prompts.

\begin{figure}[t]
    \centering
    \includegraphics[width=\linewidth]{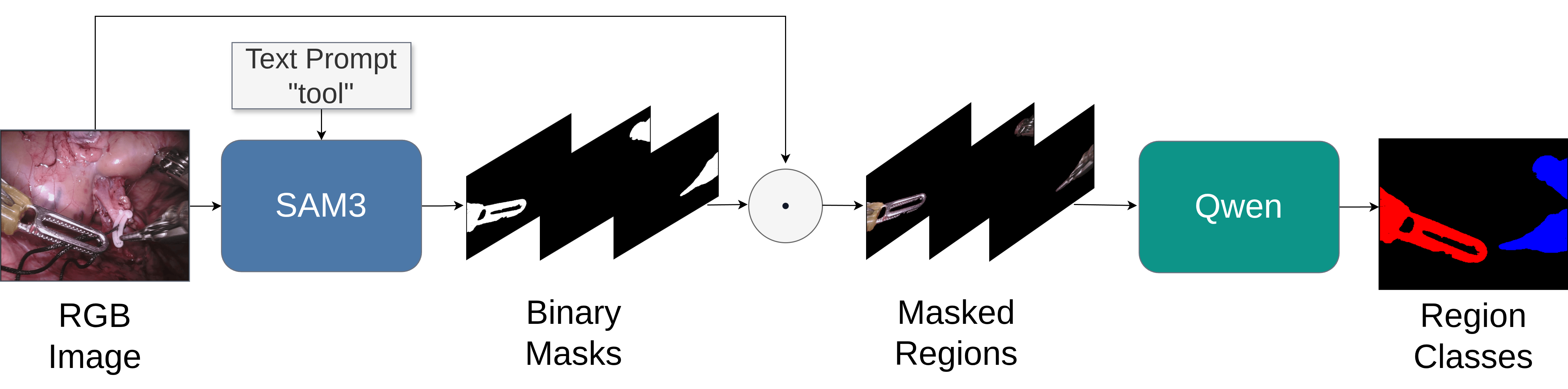}
    \caption{Overview of the proposed pipeline. SAM3 generates binary masks from the input RGB image using a fixed text prompt. These masks are applied to the image to obtain masked regions. These regions are classified by Qwen, and the predicted labels are assigned to produce the final instance-level segmentation map.}
    \label{fig:pipeline}
\end{figure}

To address these challenges while leveraging SAM3’s text-based prompting capability, we propose a two-stage framework. In the first stage, SAM3 is prompted with a fixed, natural-language-aligned text prompt, "tool", to generate instrument category agnostic binary masks. In the second stage, these masks are integrated with the original RGB frame and used as input to a Qwen~\cite{bai2025qwen2} Vision-Language Model (VLM) for instrument classification. Our choice of Qwen is motivated by our prior evaluation of general-purpose VLMs for surgical tool detection~\cite{poudel2026evaluating}, in which Qwen achieved the strongest instrument-recognition performance among the models considered. By decoupling the "where" (localization via SAM3) from the "what" (classification via Qwen), our framework provides a pipeline for instance-level surgical instrument segmentation that bypasses the pixel-level data annotation and spatial interaction, as shown in Figure~\ref{fig:pipeline}.

% refrence ~\cite{ayobi2025pixel}

% \begin{table}[t]
% \centering
% \caption{SAM3 performance on the EndoVis 2017 and 2018 datasets using point, box, and instrument category names as text prompts. Binary IoU and Instrument-wise IoU (Inst IoU) evaluate binary and instrument-level segmentation. For instrument-level segmentation, the output instrument labels are inherited from the input prompts. Results are adopted from~\cite{dong2025more}.}
% \setlength{\tabcolsep}{6pt}

% \label{tab:endovis_results}
% \begin{tabular}{lccccc}
% \toprule
% \multirow{2}{*}{Prompt} 
% & \multicolumn{2}{c}{EndoVis2017} 
% & \multicolumn{2}{c}{EndoVis2018} \\
% \cmidrule(lr){2-3} \cmidrule(lr){4-5}
% & Binary IoU & Inst IoU & Binary IoU & Inst IoU \\
% \midrule
% Point & 79.15 & 77.57 & 82.43 & 81.14 \\
% Box & 92.61 & 91.67 & 93.33 & 91.88 \\
% Text & 20.07 & 9.09 & 48.49 & 15.99 \\
% \bottomrule
% \end{tabular}
% \end{table}
\section{Related Work}
Earlier methods for surgical instrument segmentation~\cite{shvets2018automatic,ni2020pyramid,hasan2019u,jin2019incorporating} tackle the task using pixel-level classification dominated by encoder-decoder architectures, most notably UNet~\cite{ronneberger2015u} and its variants. Later,~\cite{gonzalez2020isinet,ayobi2023matis,baby2023forks,zhang2025surgical,zhao2022trasetr} introduced instance-level classification by first determining the instrument instances and adding class information to them. While these methods have demonstrated steady improvements in segmentation accuracy, they remain heavily dependent on pixel-level ground truth annotations. Recent integration of vision-language models~\cite{zhou2023text} has introduced a text-promptable mechanism to enhance generalizability, yet this method still requires mask-level supervision.

Several studies have already explored the use of SAM and SAM2 for instrument segmentation. For instance, Lou {\it et al}. \cite{lou2025zero} applied SAM2 to zero-shot segmentation; however, the approach relies on manual spatial prompting. To bypass manual input, several studies have integrated trained detectors \cite{sheng2024surgical} or introduced class-promptable mechanisms \cite{yue2024surgicalsam}. Additionally, other studies~\cite{wang2023sam,zhang2023customized} have focused on domain-specific fine-tuning on the instrument dataset. Nevertheless, these solutions necessitate either training a detector or the fine-tuning of specialized modules, both of which require ground truth annotations. In contrast, our work leverages recently released SAM3's text-based prompting capabilities, effectively eliminating the need for both the spatial prompt and pixel-level data annotation.

\section{Methodology}
\subsection{Problem Deﬁnition}

We define the input endoscopic image as $I \in \mathbb{R}^{H \times W \times 3}$. Let $p$ denote a fixed text prompt (i.e., "tool"), and let $\mathcal{G}_\phi(I, p)$ represent the segmentation model (SAM3) with pretrained parameters $\phi$. The model produces a set of candidate binary masks $\{M_k\}_{k=1}^{K}$, with $M_k \in \{0,1\}^{H \times W}$. Each mask is applied to the image via element-wise multiplication, yielding masked regions: $I_k = I \odot M_k$. A vision--language model (Qwen) $\mathcal{C}_\psi(I_k)$, where $\psi$ denotes learned parameters, assigns an instrument category $y_k \in \mathcal{C}$ to each masked region. The final instance-level segmentation map $S \in \{0,1,\dots,\mathcal{C}\}^{H \times W}$,
where $0$ denotes the background, is constructed by assigning label, $y_k$ to all pixels where  $M_k = 1$, and $0$ to all remaining pixels. The solution is therefore to produce $S$ from $I$ without spatial interaction and ground truth mask supervision.
\subsubsection{Mask Generation using SAM3:}
\begin{figure}[!t]
    \centering
    \includegraphics[width=\linewidth]{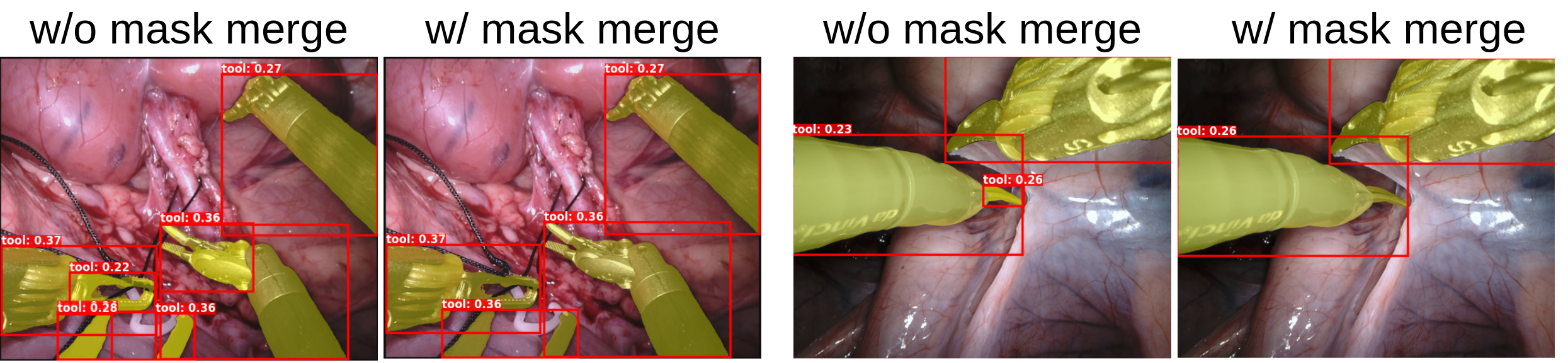}
    \caption{Illustration of segmentation outputs with and without mask merging. Overlapping masks are combined to produce a clean and complete mask. }
    \label{fig:merge}
\end{figure}

The pretrained Segment Anything Model 3 is utilized in a zero-shot manner to generate binary masks of surgical instruments. The process takes an input RGB image $I$, leverages the text-based prompting capabilities of SAM3 to obviate the need for manual interactions such as bounding boxes or point prompts, and supports automated mask generation. While instrument-specific prompts (i.e., using the instrument name as the prompt) resulted in degraded mask quality due to domain mismatch, the general prompt "tool" better aligns with the language used during SAM3 pretraining, facilitating the generation of a set of better quality binary masks, $\{M_k\}_{k=1}^{K}$. We apply a confidence threshold $t$ to filter masks, followed by an IoU-based mask merging strategy with an IoU threshold $m$ to reduce overlapping masks within the instrument. A visualization of the mask merging is shown in Figure~\ref{fig:merge}.

\subsubsection{Tool Classification with Qwen:}
% The Qwen model is finetuned using masked regions generated using SAM3 on the training set.
% To obtain instance-level segmentation, each predicted mask is applied to the original image via element-wise multiplication, producing masked RGB regions $I_k$ corresponding to candidate instruments. These regions are provided as input to a vision–language classifier (Qwen), $\mathcal{C}_\psi(I_k)$, which predicts the semantic label $y_k$ for each instance. Notably, the classifier is fine-tuned using SAM-generated masks rather than ground truth segmentation masks, thereby enabling assessment of SAM3’s intrinsic mask-generation capability without mask-level supervision. The predicted labels are subsequently assigned to their respective mask regions to construct the final semantic instance segmentation map, $S$.

% For the remainder of the manuscript, we refer to the mask images produces using SAM3 masks as \textit{SAM3-masked regions} and the mask images obtained using ground truth masks as \textit{ground truth-masked regions}.

The Qwen model, $\mathcal{C}_\psi$, is fine-tuned on the training set using masked regions derived from SAM3 predictions, following the mask generation process described above. Specifically, masked regions are obtained by applying each predicted binary mask to the original image via element-wise multiplication, yielding masked regions $I_k$ corresponding to candidate instruments. Importantly, fine-tuning Qwen with masked regions derived from SAM3-predicted masks, rather than ground truth masks, maintains a fully mask annotation-free training pipeline.

During inference, masked regions from the test set are fed to the fine-tuned Qwen model, $\mathcal{C}_\psi(I_k)$, which predicts the class label $y_k$. The predicted labels are subsequently assigned to their respective candidate instrument regions to construct the final semantic instance segmentation map, $S$.

For the remainder of the manuscript, we refer to masked regions obtained using SAM3-generated masks as \textit{SAM3 masked regions} and those obtained using ground truth masks as \textit{ground truth masked regions}.\section{Experiments}
\subsection{Dataset}
We used three datasets in the experiments, which consist of frames from robot-assisted surgery videos: EndoVis 2017~\cite{allan20192017}, EndoVis 2018~\cite{allan20202018}, and GraSP~\cite{ayobi2025pixel}. The brief descriptions of each are given below.

 The EndoVis 2017 dataset contains 1,800 annotated frames. Following the protocol in~\cite {shvets2018automatic}, we use 4-fold cross-validation, with each fold comprising $1,350$ frames for training and $450$ frames for testing. The dataset contains seven surgical instrument categories: Bipolar Forceps (BF), Prograsp Forceps (PF), Large Needle Driver (LND), Vessel Sealer (VS), Grasping Retractor (GR), Monopolar Curved Scissors (MCS), and Ultrasound Probe (UP).

For EndoVis 2018, we follow the labeling and data splits proposed in~\cite {gonzalez2020isinet}, which comprise $1,639$ training frames and $596$ test frames. The training and testing sets are drawn from different video sequences. Specifically, sequences $3$, $4$, $6$, and $7$ are used for training, while sequences $2$, $5$, $9$, and $15$ are used for testing. This dataset also contains seven instrument categories: Bipolar Forceps (BF), Prograsp Forceps (PF), Large Needle Driver (LND), Monopolar Curved Scissors (MCS), Suction Instrument (SI), Clip Applier (CA) and Ultrasound Probe (UP).
 
The GraSP dataset, sampled at $1$ frame per second, consists of $3,449$ frames.  We followed the original data split, using five video sequences for testing and the remaining frames for training, yielding $1,125$ frames for testing and $2,324$ for training. Since this dataset shares the same six instrument categories as the EndoVis 2018 dataset, we used it to evaluate Qwen's cross-dataset generalization.

As SAM3 is used in a training-free manner, the training sets are used only to fine-tune the Qwen model. All performance evaluations are conducted on the test splits.

% \subsection{Mask Generation Setup}
% Masks are generated solely from a text prompt, i.e., "tool". The predicted masks are filtered using a confidence threshold of $0.4$ and non-maximum suppression (NMS) with an IoU threshold of $0.4$. These masks are then combined with the corresponding RGB images to produce masked regions. 
\subsection{Qwen Fine-tuning and Testing Setup}
We fine-tuned the Qwen model on the SAM3-masked regions. As SAM3 can mis-segment tissue regions as instruments, we introduced an additional "No Tool" category during fine-tuning. This allows the Qwen model to act as a filter to discard mis-segmented regions. In addition to our proposed approach, we further fine-tuned Qwen on ground truth masked regions to establish an upper bound for classification performance.

Qwen evaluation was performed using ground truth masked regions from the test set. Even though all ground truth masked regions correspond to valid instruments, the classifier trained on SAM3 masked regions may occasionally predict the "No Tool" class during evaluation. To ensure fair comparison across different evaluation settings, predictions
corresponding to the "No Tool" class were replaced by randomly assigning one
of the valid instrument classes ($1$--$7$), and the results were
averaged over five trials. Since the EndoVis 2018 and GraSP datasets share the same instrument categories, we also evaluated cross-dataset generalization by fine-tuning the Qwen model on GraSP ground truth-masked regions and testing it on the EndoVis 2018 dataset.

 \subsection{Segmentation Evaluation Setup}
We evaluated both binary and instance-level segmentation performance. Binary segmentation performance was evaluated directly using SAM3-predicted masks. For instance-level evaluation, each SAM3 mask was assigned the class label predicted by Qwen, yielding an instance-level segmentation map.

Instance-level segmentation metrics heavily penalize predictions in cases of incorrect classification, even when the mask quality is near perfect. To estimate the upper-bound instance segmentation performance achievable with SAM3 masks, we isolate the effect of classification errors by considering an oracle classifier that assigns the ground truth class label to each predicted mask based on its maximum spatial overlap with the ground truth masks. Because SAM3 may again mis-segment non-instrument regions, these regions were assigned a random class label between $1$ and $7$. The final results were averaged over five trials. While the oracle evaluation removes classification error, it still preserves errors arising from imperfect mask generation, including missed detections, inaccurate segmentation boundaries, and overlapping segmentations.

Furthermore, for comparison, we evaluate the original SAM3 as a baseline for both binary and instance segmentation using instrument-specific text prompts. For each frame, prompts are constructed using the names of instruments present in the scene, obtained from ground truth class labels. Predicted masks are assigned class labels based on their corresponding prompts. In cases where a pixel is covered by multiple masks, the label corresponding to the highest-confidence SAM3 prediction is assigned.

\subsection{Implementation}
We utilized the original pretrained SAM3 model for mask generation. The predicted masks were filtered using a confidence threshold of $t=0.2$ and merged using an IoU threshold of $m=0.02$. The Qwen 2.5-7B model was fine-tuned using supervised Rank-8 LoRA~\cite{hu2022lora} adaptation for $5$ epochs, with a batch size of $4$, a learning rate of $1\times10^{-4}$, and a gradient accumulation step of $4$. Training was performed using the Swift\footnote[6]{\url{https://github.com/modelscope/ms-swift}} framework, optimized with the Adam optimizer, and employed the standard token-level cross-entropy (next-token prediction) loss. All experiments were conducted on an NVIDIA A100 GPU~\cite{https://doi.org/10.34788/0s3g-qd15}.
\subsection{Evaluation}
We used Binary IoU and Dice score to evaluate binary segmentation performance, and three IoU-based metrics -- Ch\_IoU, ISI\_IoU, and mc\_IoU -- from~\cite{gonzalez2020isinet} to assess instance segmentation performance. The Ch\_IoU evaluates performance considering only the instrument classes present in the ground truth mask. In contrast, the ISI\_IoU considers classes across both ground truth and predictions, and therefore penalizes additional incorrect classes predicted by the model. Both metrics are computed per frame and averaged over all frames.
The mean class IoU (mc\_IoU) computes the IoU for each class across all frames and then averages over classes, ensuring equal contribution from each class. To separately evaluate the classification performance of Qwen, we report Accuracy and macro F1-score.

% \begin{table*}[t]
% \centering
% \caption{Classification performance of Qwen fine-tuned with different masked regions.}
% \label{tab:cls_results}
% \begin{tabular}{lccc|ccc}
% \toprule
% \multirow{2}{*}{Training Mask} 
% & \multicolumn{3}{c|}{EndoVis 2018} 
% & \multicolumn{3}{c}{GraSP} \\
% \cmidrule(lr){2-4} \cmidrule(lr){5-7}
% & Accuracy & Macro F1 & Wt F1 
% & Accuracy & Macro F1 & Wt F1 \\
% \midrule
% SAM3 mask & $72.00$ & $0.40$ & $0.70$ & $86.74$ & $0.81$ & $0.86$ \\
% GraSP GT mask & $76.09$ & $0.52$ & $0.78$ & -- & -- & -- \\
% GT mask   & $76.67$ & $0.53$ & $0.76$ & $89.79$ & $0.85$ & $0.90$ \\
% \bottomrule
% \end{tabular}
% \end{table*}

% \begin{table}[h]
% \centering
% \caption{Classification performance of Qwen fine-tuned with different masked regions.}
% \label{tab:cls_results}
% \begin{tabular}{lccc|ccc}
% \toprule
% \multirow{2}{*}{Training Mask} 
% & \multicolumn{3}{c|}{EndoVis 2017} 
% & \multicolumn{3}{c}{EndoVis 2018} \\
% \cmidrule(lr){2-4} \cmidrule(lr){5-7}
% & Accuracy & Macro F1 & Wt F1 
% & Accuracy & Macro F1 & Wt F1 \\
% \midrule
% SAM3 mask & $0.67\pm0.10$ & $0.44\pm0.11$ & $0.67\pm0.11$ & $0.72$ & $0.40$ & $0.70$  \\
% GraSP GT mask & -- & -- & --& $0.76$ & $0.52$ & $0.78$  \\
% GT mask  & $0.80\pm0.13$ & $0.61\pm0.38$ & $0.81\pm0.13$ & $0.76$ & $0.53$ & $0.76$  \\
% \bottomrule
% \end{tabular}
% \end{table}

\begin{table}[h]
\centering
\caption{Classification performance of Qwen fine-tuned with different masked
regions across three different datasets. Evaluation is performed on ground truth masked regions, and performance is reported in terms of accuracy and macro F1-score. $\uparrow$ indicates that higher values are desired.}
\setlength{\tabcolsep}{6pt}

\label{tab:cls}
\begin{tabular}{lcc|cc|cc}
\hline
 & \multicolumn{2}{c}{EndoVis 2017} 
 & \multicolumn{2}{c}{EndoVis 2018}
 & \multicolumn{2}{c}{GraSP} \\
Fine-tuning 
& Acc $\uparrow$  & Macro F1 $\uparrow$ 
& Acc $\uparrow$  & Macro F1 $\uparrow$
& Acc $\uparrow$ & Macro F1 $\uparrow$ \\
\hline
SAM3 & $62.76$ & $40.30$ & $72.33$ & $43.70$ & $86.08$ & $80.31$  \\
GT   & $74.33$ & $48.75$ & $76.67$ & $53.05$  & $89.79$ & $85.16$ \\
GraSP GT & $-$ & $-$ & $76.09$ & $52.44$ & $-$ & $-$ \\
% SAM3/SAM3 & $-$ & $-$ &  &  & $-$ & $-$ \\
% SAM3/GT & $-$ & $-$ & $76.09$ & $52.44$ & $-$ & $-$ \\

% Method E &  &  &  &  &  &  \\
% Method F &  &  &  &  &  &  \\
\hline
\end{tabular}
\end{table}
\section{Results}

% \begin{table}[t]
% \centering
% \caption{Binary segmentation performance comparison across datasets.}
% \label{tab:binary_all}
% \begin{tabular}{lcc|cc|cc}
% \hline
%  & \multicolumn{2}{c}{EndoVis 2017} 
%  & \multicolumn{2}{c}{EndoVis 2018}
%  & \multicolumn{2}{c}{GraSP} \\
% Method & IoU & Dice & IoU & Dice & IoU & Dice \\
% \hline
% UNet & $75.44$ & 84.37 & 68.89  &  --  &  --  &  -- \\
% % TernausNet & 83.60 & 90.01 &  --  &  --  &  --  &  -- \\
% % MF-TAPNet  & 87.56 & 93.37 &  --  &  --  &  --  &  -- \\\hline
% SAM3     & $74.83$ & $82.10$ & $83.70$ & $88.73$ & 81.58 & xx.xx \\
% \hline
% \end{tabular}
% \end{table}

\subsection{Qwen Classification Performance}
We evaluated the classification performance of the Qwen model separately using accuracy and the macro F1-score as the evaluation metrics. The results are summarized in Table~\ref{tab:cls}.  

While fine-tuning on SAM3-masked regions leads to notable performance degradation on the EndoVis 2017 dataset, the reduction in accuracy on EndoVis 2018 is relatively low. However, the macro F1-score on the latter decreases substantially, due to a large performance reduction in the minority classes. Performance is highest on the GraSP dataset, with only a small reduction in both accuracy and macro F1-score, likely aided by the larger number of samples available for fine-tuning. Furthermore, the cross-dataset evaluation shows that a Qwen model fine-tuned on the GraSP dataset achieves performance on EndoVis 2018 nearly equivalent to that of fine-tuning on the ground truth, highlighting Qwen's strong cross-dataset generalization.

% The performance of the Qwen classifier was evaluated on RGB masks extracted from ground truth annotations. Two training strategies were considered to analyze the impact of mask quality. First, the model was fine-tuned using RGB masks generated from SAM3 outputs. Since SAM3 occasionally over-segments surrounding tissues as instruments, an additional class, “No Tool”, was introduced to account for such mis-segmentations. Second, the model was fine-tuned using RGB masks extracted directly from ground truth segmentation masks to compare the effect of training with SAM3-generated masks versus ground truth masks.

% Furthermore, since the EndoVis 2018 and GraSP datasets share the same instrument categories, we evaluated cross-dataset generalization by testing the Qwen model fine-tuned on GraSP RGB masks (ground truth) on the EndoVis 2018 dataset. However, because the “No Tool” class is present only in the SAM3-generated training masks, the classifier may occasionally predict this class during validation even though ground truth masks always correspond to valid instruments. To ensure fair comparison across different training settings, predictions corresponding to the “No Tool” class were replaced by randomly assigning one of the valid instrument classes (1–7) during evaluation, and the results were averaged over five trials.
\begin{table}[h]
\centering
\setlength{\tabcolsep}{5pt}
\caption{Binary segmentation performance of UNet and SAM3 using instrument-specific prompt and "tool" as a prompt (highlighted) on the EndoVis 2017 and EndoVis 2018 datasets. Performance is evaluated using Binary Intersection-over-Union (IoU) and Dice score. $\uparrow$ indicates that higher values are desired.}
\label{tab:binary_all}
\begin{tabular}{lcc|cc}
\hline
 & \multicolumn{2}{c}{EndoVis 2017} 
 & \multicolumn{2}{c}{EndoVis 2018} \\
Method & Binary IoU $\uparrow$& Dice $\uparrow$ &Binary IoU $\uparrow$ & Dice $\uparrow$ \\
\hline 
UNet~\cite{ronneberger2015u}      & $75.44 \pm 18.18$ & $84.37 \pm 14.58$ & --  &  --  \\
% TernausNet & 83.60 & 90.01 &  --  &  --  \\
% MF-TAPNet  & 87.56 & 93.37 &  --  &  --  \\
\hline
SAM3\textsubscript{inst}(Baseline) & $16.51 \pm 31.21$ & $19.20 \pm 34.41$ & $12.26 \pm 26.19$ & $15.13 \pm 29.13$ \\
\rowcolor{green!15}
SAM3\textsubscript{tool} & $74.12 \pm 31.47$ & $82.10 \pm 23.23$ & $79.75 \pm 27.40$ & $84.93 \pm 27.06$ \\
\hline
\end{tabular}
\end{table}

\subsection{Binary Segmentation}
The binary segmentation performance on the EndoVis 2017 and EndoVis 2018 datasets is summarized in Table~\ref{tab:binary_all}. We compare a vanilla UNet with SAM3 on EndoVis 2017. Metrics values for UNet on EndoVis 2018 were not available. Performance is evaluated using Intersection-over-Union (IoU) and Dice score.

Our approach, using SAM3 with "tool" as the text prompt, achieved performance comparable to UNet, a fully supervised method, on the EndoVis 2017 dataset. In contrast, using instrument-specific prompts resulted in a substantial drop in performance. Similar performance is also observed on the EndoVis 2018 dataset. 

% On EndoVis 2017, U-Net achieves an IoU of $75.44 \pm 18.18$ and a Dice score of $84.37 \pm 14.58$, while SAM3 attains a comparable IoU of $74.83 \pm 26.67$ and Dice score of $82.10 \pm 23.23$. On EndoVis 2018, SAM3 achieves an IoU of $83.70 \pm 22.29$ and Dice of $88.73 \pm 20.87$. These results highlight SAM3's strong zero-shot binary segmentation performance.

\begin{table}[!t]
\centering
\caption{Instance-level segmentation performance on the EndoVis 2017 dataset. The upper section lists several supervised methods, while the lower section reports SAM3 with instrument-specific prompts as a baseline, our approach (highlighted), and SAM3 with an oracle classifier. Three metrics, Ch\_IoU, ISI\_IoU, and mc\_IoU, are used to evaluate performance. The instrument names are abbreviated.}
\label{tab:instance_endovis}
% \begin{subtable}{\textwidth}
% \caption{EndoVis 2017}
\label{tab:instance_endovis2017}
\setlength{\tabcolsep}{2pt}
\resizebox{\textwidth}{!}{
\begin{tabular}{l|c|c|ccccccc|c}
\toprule
Method 
& Ch\_IoU 
& ISI\_IoU 
& BF 
& PF 
& LND 
& VS  
& GR 
& MCS  
& UP 
& mc\_IoU \\
\midrule
TernausNet~\cite{shvets2018automatic} & $35.27$ & $12.67$ &$13.45$  & $12.39$ & $20.51$  & $5.97$ & $1.08$ & $1.00$ & $16.76$ & $10.17$ \\
MF-TAPNet~\cite{jin2019incorporating} & $37.25$  & $13.49$ & $16.39$ & $14.11$ &  $19.01$ & $8.11$  & $0.31$ & $4.09$ & $13.40$ & $10.77$ \\
ISINet~\cite{gonzalez2020isinet} & $55.62$ & $52.20$ & $38.70$ & $38.50$  & $50.09$ & $27.43$ & $2.10$ & $28.72$ & $12.56$ & $28.96$ \\
TraSeTR~\cite{zhao2022trasetr} & $60.40$ & -- & $45.20$ & $56.70$ & $55.80$ & $38.90$ & $11.40$ & $31.30$ & $18.20$ & $36.79$ \\
S3Net~\cite{baby2023forks} & $72.54$ & $71.99$ & $75.08$ & $54.32$ & $61.84$ & $35.50$ & $27.47$ & $43.23$ & $28.38$ & $46.55$ \\
MATIS~\cite{ayobi2023matis} & $66.73$ & $61.79$ & $67.17$ & $50.36$ & $46.53$ & $31.49$ & $11.08$ & $13.57$ & $24.79$ & $34.99$ \\
Zhang {\it et al}.~\cite{zhang2025surgical} & $69.79$ &$64.78$  & $61.03$ & $52.28$  & $45.69$ & $34.66$ & $15.00$ & $20.81$ & $27.14$ & $36.66$ \\
\midrule
SAM3\textsubscript{inst}(Baseline)    & $4.23$ & $4.23$ &$1.21$  & $13.37$ & $0.00$ & $0.00$ & $8.64$ & $0.00$ & $0.11$ & $3.33$ \\
\rowcolor{green!15}
SAM3\textsubscript{tool}+Qwen & $40.56$ & $35.07$ & $40.48$ & $24.61$ & $25.81$ & $25.50$ & $7.47$ & $21.74$ & $7.04$ & $21.81$ \\

% \midrule

SAM3\textsubscript{tool}+Oracle   & $61.49$ & $59.97$ & $70.45$ & $54.25$ & $38.43$ & $36.41$ & $33.96$ & $39.96$ & $22.62$ & $42.30$ \\

% SAM 3 + ResNet  &  &  &  &  &  &  &  &  &  &  \\
\bottomrule
\end{tabular}
}
\end{table}

\begin{table}[h]
\centering
\caption{Instance-level segmentation performance on the EndoVis 2018 dataset. The upper section lists several supervised methods, while the lower section reports SAM3 with instrument-specific prompts as a baseline, our approach (highlighted), and SAM3 with an oracle classifier. Three metrics, Ch\_IoU, ISI\_IoU, and mc\_IoU, are used to evaluate performance. The instrument names are abbreviated.}
\label{tab:instance_endovis2018}
\setlength{\tabcolsep}{2pt}
\resizebox{\textwidth}{!}{
\begin{tabular}{l|c|c|ccccccc|c}
\toprule
Method 
& Ch\_IoU 
& ISI\_IoU 
& BF 
& PF 
& LND 
& MCS 
& SI 
& CA 
& UP 
& mc\_IoU \\
\midrule
TernausNet~\cite{shvets2018automatic} & $46.22$ & $39.87$ & $44.20$  & $4.67$  & $0.00$  & $0.00$ & $0.00$ & $50.44$ & $0.00$ & $14.19$ \\
MF-TAPNet~\cite{jin2019incorporating} & $67.87$  & $39.14$ & $69.23$ & $6.10$ & $11.68$ & $14.00$ & $0.91$ & $70.24$ & $0.57$ & $24.68$ \\
ISINet~\cite{gonzalez2020isinet} & $73.03$ & $70.94$ & $73.83$ & $48.61$ & $30.98$ & $37.68$ & $0.00$ & $88.16$ & $2.16$ & $40.21$ \\
TraSeTR~\cite{zhao2022trasetr} & $76.20$ & -- & $76.30$ & $53.30$ & $46.50$  & $40.60$ & $13.90$ & $86.20$  & $2.16$ & $47.71$ \\
S3Net~\cite{baby2023forks} & $75.81$ & $74.02$ & $77.22$ & $50.87$ & $19.83$ & $50.59$ & $0.00$  & $92.19$ & $7.44$ & $42.58$ \\
MATIS~\cite{ayobi2023matis} & $82.31$ & $77.01$ & $83.55$ & $38.65$ & $40.48$ & $92.56$ & $70.38$ & $0.00$ & $14.39$ & $48.57$ \\
Zhang {\it et al}.~\cite{zhang2025surgical} & $84.67$ &$79.86$  & $83.95$ & $41.47$ & $66.57$  & $92.74$ & $74.20$ & $0.00$ & $23.50$ & $54.63$ \\
\midrule
% SAM3 (Baseline)    & $-$ & $-$ &$-$  & $-$ & $-$ & $-$ & $-$ & $0.00$ & $-$ & $-$ \\

SAM3\textsubscript{inst}(Baseline)& $7.79$ & $7.79$ &$15.66$  & $4.74$ & $0.00$ & $0.00$ & $0.00$ & $0.00$ & $0.00$ & $2.92$ \\
\rowcolor{green!15}
SAM3\textsubscript{tool}+Qwen & $61.53$ & $58.80$ &$69.69$  & $18.03$ & $14.83$ & $73.49$ & $27.18$ & $2.92$ & $3.55$ & $29.96$ \\

% \midrule
SAM3\textsubscript{tool}+Oracle& $79.39$ & $76.66$ & $80.12$ & $51.35$ &$71.34$  & $83.95$ & $64.25$  & $45.90$ & $19.65$ & $59.51$ \\

% SAM 3 + ResNet  &  &  &  &  &  &  &  &  &  &  \\
\bottomrule
\end{tabular}
}
\end{table}

\subsection{Instance Segmentation}

The instance-level segmentation performance is reported in Table~\ref{tab:instance_endovis2017} for the EndoVis 2017 dataset and in Table~\ref{tab:instance_endovis2018} for the EndoVis 2018 dataset. The upper section of the table lists several state-of-the-art supervised methods, while the lower section includes SAM3 with instrument-specific prompts as a baseline and our approach with Qwen classifier and SAM3 with an oracle classifier. Performance is evaluated using three IoU-based metrics: Ch\_IoU, ISI\_IoU, and mc\_IoU.

While supervised models inherently maintain a performance advantage, our objective at this stage is to challenge the necessity of the manual mask-annotating bottleneck. Despite operating in an entirely mask annotation-free setting, our framework manages to outperform established benchmarks like TernausNet and MF-TAPNet across both EndoVis 2017 and 2018 datasets. This highlights the potential of our framework in surgical environments where labeled data is scarce.

Compared with directly leveraging SAM3 with an instrument-specific prompt, our approach achieved significantly better performance. The baseline completely failed to handle some instruments with $0$ IoU, whereas our approach handles diverse instrument categories.

% While the proposed SAM3 + Qwen  framework currently lags behind recent fully supervised state-of-the-art methods, it is imperative to evaluate these results within an annotation-free paradigm.

% Our approach shows significant performance gains compared to the original instrument-specific prompting. As reported in Table~\ref{tab:endovis_results}, the Ch\_IoU for EndoVis 2017 using standard SAM3 instrument-category text prompts was as low as $9.09$. With our approach, this metric has improved to $42.93$ (Table~\ref{tab:instance_endovis2017}). Similar substantial improvements were observed across the EndoVis 2018 dataset.

Furthermore, oracle analysis yields performance comparable to many supervised methods. Since the oracle represents a theoretical upper bound, such a classifier may not be realizable in practice. This, however, indicates that further improvements in classifier performance could directly enhance segmentation results, given the underlying mask quality produced by SAM3. 

\begin{wrapfigure}[16]{r}{0.45\textwidth}
\centering
    \includegraphics[width=0.44\textwidth]{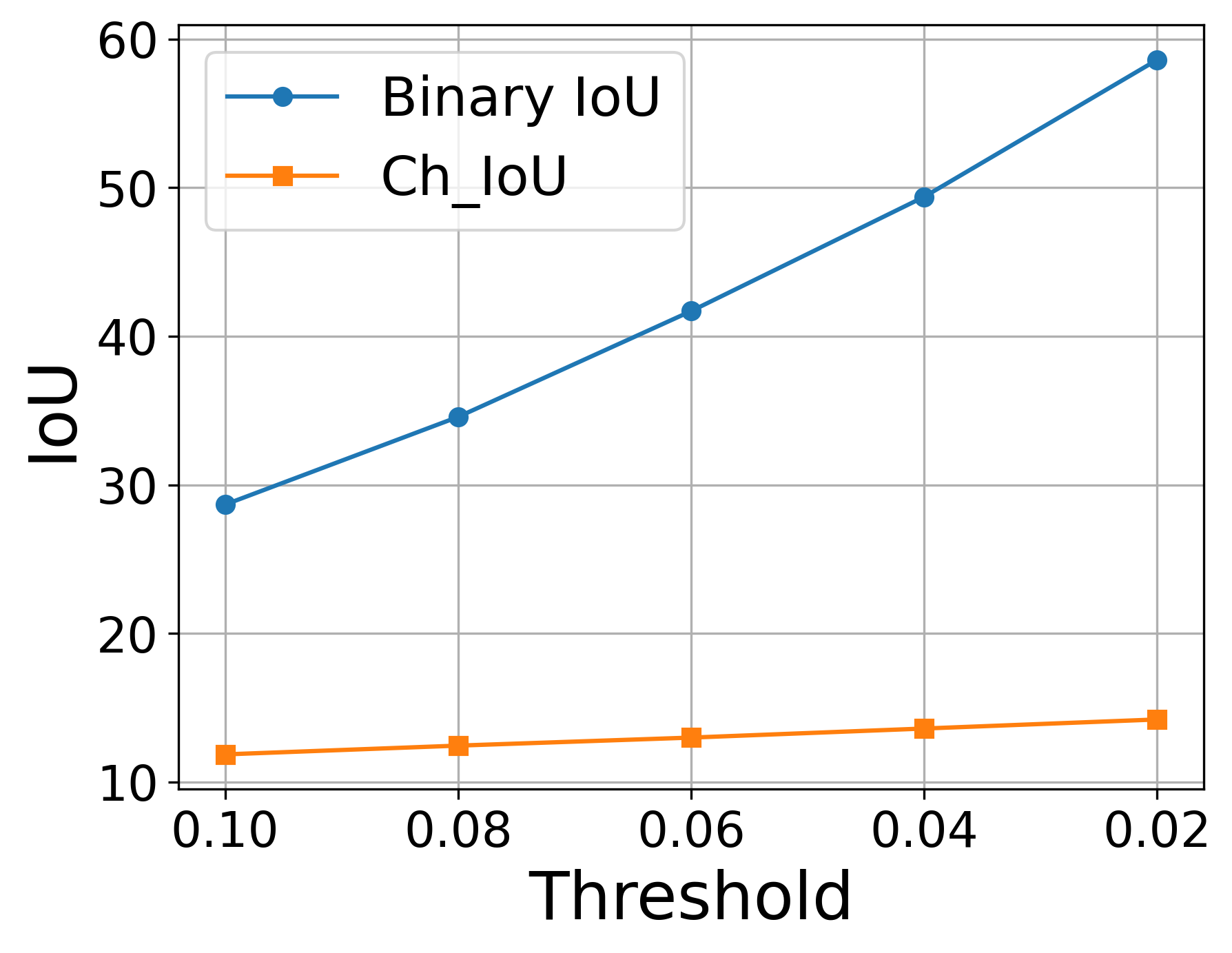} % adjust width a bit smaller than wrap width
    \captionsetup{font=small}
    \caption{Plot of IoU vs. confidence threshold for the baseline model on EndoVis 2018 dataset.}
    \label{fig:sidefigure}
\end{wrapfigure}

We observe that using SAM3 baseline with the same confidence threshold $t$  results in many missing predictions. Therefore, we experimented with lowering the segmentation confidence threshold. The plot of IoU versus threshold is shown in Figure~\ref{fig:sidefigure} for the EndoVis 2018 dataset. While lowering the threshold increases binary segmentation performance by producing more predictions, it fails to correctly identify the instrument categories, resulting in poor instance-level segmentation performance. Additionally, lower thresholds result in poor mask quality, characterized by many fragmented, overlapping, and redundant predictions.

% When comparing with SA2's original instrument-specific prompting, our approach shows significant performance gains. For example, the Ch_IoU and ISI_IoU for EndoVis 2017 using SAM3 instrument-category text prompt was $9.09$ (Table 1), our approach has improved it to $66.76$. Similar improvements is also obtained for Endovis 2018 dataset.  Furthermore, oracle analysis reveals identical performance to the baseline. However, because an oracle is a theoretical measure, such a classifier may not exist in practice. This suggests that improving classifier performance could boost segmentation performance, given the existing quality of the SAM3 masks.    

\begin{figure}[!h]
    \centering
    \includegraphics[width=\linewidth]{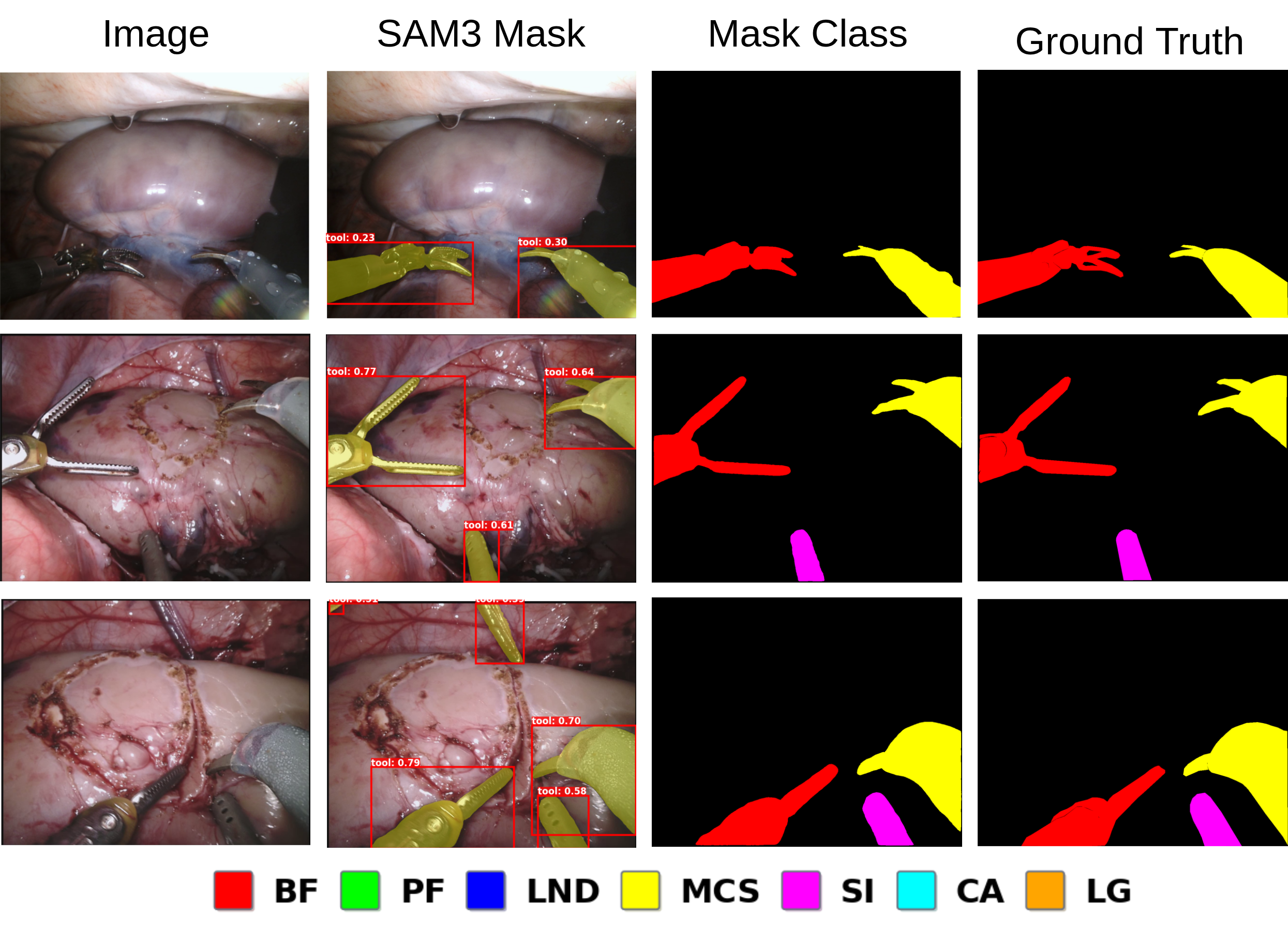}
    \caption{Instance-level segmentation of surgical instruments. The first column shows the RGB input images, the second shows masks generated by SAM3, the third shows the classified masks from Qwen, and the fourth shows the ground truth masks. Mask colors indicate instrument categories, as specified in the index with their shorthand notations.}
    \label{fig:results}
\end{figure}

The visualization of the instance-level segmentation mask using SAM3 and Qwen is shown in Figure~\ref{fig:results}. The progression from left to right illustrates the RGB input image, the initial masks generated by SAM3, the class predictions assigned by Qwen, and the corresponding ground truth annotations. The first two rows demonstrate SAM3's ability to generate tight bounding boxes and precise spatial boundaries, which are then accurately classified by Qwen. Furthermore, the third row highlights Qwen's handling of false positives: a minor over-segmentation error by SAM3 -- where small background regions are incorrectly masked as instruments -- is successfully recognized and rejected by the Qwen classifier.
\begin{figure}[!t]
    \centering
    \includegraphics[width=\linewidth]{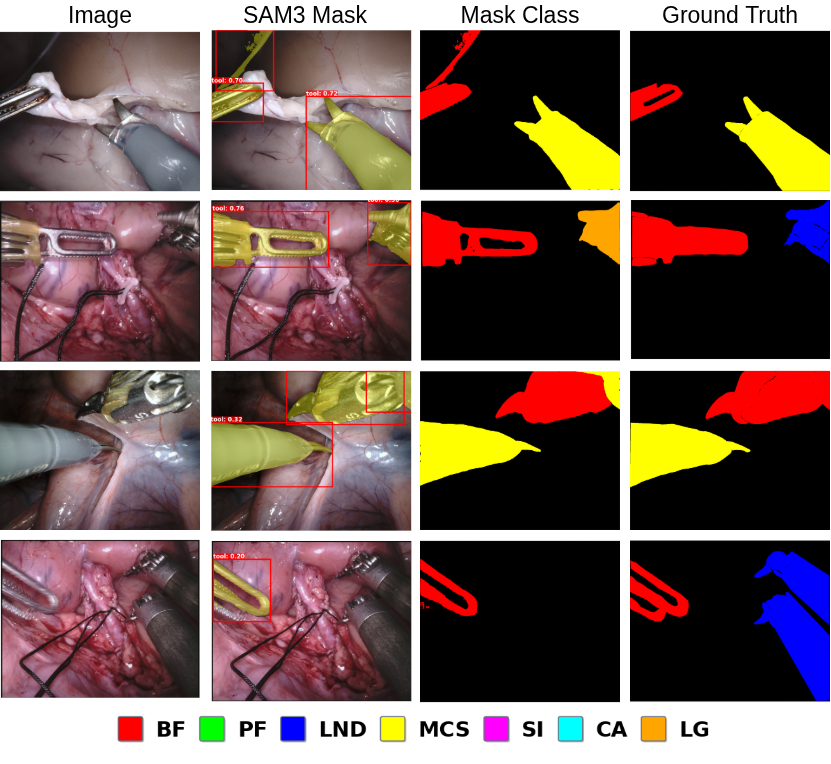}
    \caption{Examples of failure cases in instance-level segmentation of surgical instruments. The first column shows the RGB input images, the second shows masks generated by SAM3, the third shows the classified masks from Qwen, and the fourth shows the ground truth masks. Mask colors indicate instrument categories, as specified in the index with their shorthand notations.}
    \label{fig:failure}
\end{figure}

The failure cases of instance segmentation are presented in Figure~\ref{fig:failure}. The first row demonstrates a false-positive generation by SAM3, which the classifier subsequently mislabels as Monopolar Curved Scissors. The second and third rows demonstrate two distinct types of mask generation errors. The second row exhibits the overlapping detections. In this case, two large needle drivers in close proximity are segmented as a single instrument. Consequently, Qwen misclassifies the merged region as a Laparoscopic Grasper instead of Large Needle Driver. The third row highlights an issue of mask fragmentation, in which a single instrument is erroneously split into two disjoint parts. Passing these fragmented regions to the classifier leads to label inconsistency: one fragment of a Bipolar Forceps is misclassified as Monopolar Curved Scissors, while the other is correctly identified. Finally, the fourth row presents a false-negative case in which SAM3 fails to detect instruments. Ultimately, these propagated errors from the SAM3 initial localization stage limit overall performance.

% \begin{table*}[t]
% \centering
% \caption{Instance-level segmentation performance on GraSP dataset.}
% \label{tab:instance_grasp}
% \resizebox{0.6\textwidth}{!}{
% \begin{tabular}{lcc|ccccccc|c}
% \toprule
% Method 
% & Ch\_IoU 
% & ISI\_IoU 
% & mc\_IoU \\
% \midrule
% SlowFast & $77.16$  & $72.26$  & $58.75$ \\
% TAPIS &  $86.61$ & $83.38$ & $77.42$ \\

% \midrule
% SAM 3 + Qwen    & $65.47$  & $61.35$ & $48.81$ \\
% SAM 3 + Oracle   & $75.90$   &  $72.38$& $62.34$ \\

% % SAM 3 + ResNet  &  &  &  \\
% \bottomrule
% \end{tabular}
% }
% \end{table*}

\section{Discussion}
The generation of precise ground truth segmentation masks that can be subsequently used to train various deep learning-based segmentation and classification models is challenging because it requires expert knowledge and incurs significant time and cost.
To address these challenges, we aim to eliminate the need for expert annotated masks used for training and therefore propose a two-stage mask annotation-free pipeline that integrates the Segment Anything Model 3 with a Qwen model. In the first stage, SAM3 is prompted with a natural-language query (i.e.,  "tool") to generate binary masks. In the second stage, a Qwen classifier -- fine-tuned on SAM3-generated masked regions -- predicts the instrument label corresponding to each mask.

Existing methods for surgical instrument segmentation rely on either ground truth annotations or manual spatial prompts, thereby limiting scalability and automation. While SAM3 enables zero-shot segmentation via text-based prompting, its pretraining on natural image–language data introduces a domain mismatch when directly using instrument category names as prompts. To mitigate this, we employ a natural-language-aligned prompt to generate instrument category agnostic masks, improving alignment with the model’s prior while decoupling mask generation from classification. This design necessitates an additional classification stage to assign semantic labels to the generated regions.

To this end, we incorporate a Qwen vision–language model for classification, a choice grounded in our prior systematic evaluation of general-purpose VLMs for surgical tool detection~\cite{poudel2026evaluating}, where Qwen consistently outperformed alternatives such as LLaVA and InternVL in instrument recognition. We further validate this design choice by fine-tuning Qwen on both ground truth-masked and SAM3-masked regions, observing a small reduction in accuracy, particularly on large datasets such as EndoVis 2018 and GraSP. Furthermore, its strong cross-dataset generalization supports its suitability. However, despite this generalization capability, the need for re-training when new instrument classes are introduced remains a limitation and may hinder scalability in dynamic surgical environments.

% To this end, we incorporate a Qwen vision–language model for classification. We validate this design choice by fine-tuning Qwen on both ground truth-masked and SAM3-masked regions, observing a small reduction in accuracy, particularly on large datasets such as EndoVis 2018 and GraSP. Furthermore, its strong cross-dataset generalization supports its suitability. However, despite this generalization capability, the need for re-training when new instrument classes are introduced remains a limitation and may hinder scalability in dynamic surgical environments.

Additionally, the framework's overall performance is sensitive to the quality of the SAM3-generated masks. Fragmented, duplicate, and incorrect segmentations are common in challenging surgical scenarios, which degrade mask quality in the early stage, then propagating the errors to the classification stage, leading to reduced performance.

Compared with recent fully supervised approaches, the proposed framework is far from matching the performance. However, it is important to emphasize that our primary objective at this stage is not to surpass supervised models -- which rely on high-quality manual annotations -- but to reduce dependence on costly data annotation.
The ability to perform instance segmentation without ground truth masks or manual prompts represents a promising step toward its applicability in surgical environments, where annotation resources are limited.

Future work will focus on improving mask quality through annotation-free adaptation strategies, such as prompt optimization and mask refinement~\cite{lin2025samrefiner}. Additionally, leveraging medical vision–language models~\cite{jiang2025hulu} for zero-shot classification may further reduce the need for fine-tuning on noisy input data. Finally, extending the framework to real-time applications~\cite{ceron2022real} and video-based segmentation~\cite{zhang2025surgical} remains an important direction for clinical integration.

\section{Conclusion}

In this paper, we introduced a framework for instance-level surgical instrument segmentation that incorporates the SAM3 and the Qwen model. Unlike prior approaches, the proposed method eliminates the need for manual spatial prompts and ground truth mask annotations. This work establishes a promising direction for future research to reduce dependence on data annotations while maintaining performance.

Future work will investigate surgical vision–language models for zero-shot classification and mask refinement to further enhance mask quality.

\begin{credits}
\subsubsection{\ackname} We gratefully acknowledge the support of this work by the National Institutes of Health – National Institute of General Medical Sciences under Award No. R35GM128877 and the National Science Foundation – Division of Chemical, Bioengineering and Transport Systems under Award No. 2245152. We also thank the Research Computing team at the Rochester Institute of Technology~\cite{https://doi.org/10.34788/0s3g-qd15} for providing the computing resources for this research.

\end{credits}

\bibliographystyle{splncs04}
\bibliography{refs}

\end{document}